\documentclass[10pt, a4paper]{article}

\usepackage[final]{lrec2026} 

\graphicspath{{Fig/}}
\usepackage{graphicx}
\usepackage[table]{xcolor}
\usepackage{multirow}
\usepackage{float}
\usepackage{listings}

\usepackage[skip=4pt]{caption}

\definecolor{colorLightBlue}{HTML}{DAE9F8}
\definecolor{colorMediumBlue}{HTML}{A6C9EC}
\definecolor{colorBlue}{HTML}{4D93D9}
\definecolor{colorDarkBlue}{HTML}{215C98}

\definecolor{colorLightOrange}{HTML}{FBE2D5}
\definecolor{colorMediumOrange}{HTML}{F7C7AC}
\definecolor{colorDarkOrange}{HTML}{F1A983}

\definecolor{colorBrown}{HTML}{BE4F15}

\definecolor{colorLightPurple}{HTML}{F2CEEF}
\definecolor{colorDarkPurple}{HTML}{E49EDD}

\definecolor{colorLightGreen}{HTML}{C1F0C7}
\definecolor{colorDarkGreen}{HTML}{47D359}

\definecolor{colorYellow}{HTML}{F9C002}

\definecolor{colorLightGray}{HTML}{F2F2F2}
\definecolor{colorDarkGray}{HTML}{D0D0D0}

\title{The Shape of Testimony: A Scalable Framework for Oral History Archive Comparison}

\name{Itamar Trainin, Renana Keydar, Amit Pinchevski} 

\address{Hebrew University of Jerusalem \\
         \{itamar.trainin, renana.keydar, amitpi\}@mail.huji.ac.il}

\abstract{
Researchers in Holocaust studies have often distinguished between two styles of oral survivor testimony: the USC Shoah Foundation’s interviews tend to follow a structured, interviewer-guided format, whereas the Yale Fortunoff Video Archive generally favors a more free‑form, open‑ended style. This distinction has influenced both scholarly research and the development of later archives. In this study, we critically examine that claim by conducting a large-scale computational analysis of more than 1,600 testimonies from both collections.
Leveraging discourse segmentation, topic modeling, and large language model (LLM) based analysis, we quantify the ``structuredness'' level of testimonies through topic coherence, interviewer–survivor dynamics, and the distribution of question types. Our results generally corroborate the structural differences identified in earlier research, while also revealing significant overlaps between the collections, both within individual interviews and across common narrative patterns. This complicates the simple “structured vs. free‑form” dichotomy often applied to these oral histories.
Beyond revisiting a foundational claim in Holocaust studies, our work provides a scalable, replicable framework for comparative corpus analysis. As a proof of concept, it suggests broader applications for digital oral history, narrative analysis, and the design of citizen‑science annotation platforms.
 \\ \newline  }

\begin{document}

\maketitleabstract


\section{Introduction}
A common claim in the field of Holocaust Studies suggests that the two main oral history video testimony archives, the Yale Fortunoff Video Archive for Holocaust Testimonies (Henceforth: Yale or Fortunoff) and the USC Shoah Foundation (Henceforth: USC or Shoah Foundation), present two opposite styles of interviews. It is said that the interview practices by Yale tend to be loose and open-ended, whereas USC follows a stricter format. Several studies have ascertained this difference by exploring the institutional agenda and policy as well as through close readings of selected testimonies \citep{wieviorka2006era, shenker2015reframing, pollin-galay2018ecologies} and recently also by a small-scale empirical study \citep{presner2024ethics}.

Yet to empirically evaluate these assumptions, a large-scale comparative analysis is required, which is the purpose of this paper. This analysis calls for advanced computational tools, such as those developed recently in the digital humanities and computational history, for examining large narrative corpora. To date, applying these tools to oral testimonies has been limited due to the interpretive challenge and ethical complexities involved.

In what follows, we seek to ascertain the reported difference between the two testimony styles by means of a computational comparison of over 1600 Holocaust survivor testimonies drawn from the USC and Yale archives. Using large language models (LLMs), dialogical segmentation, topic modeling, and question classification, we develop a reproducible pipeline to measure ``structuredness'' across interviews. We examine topical coherence, interviewer-survivor dynamics, and the evolution of question types along interviews to assess whether and how these archives actually diverge.

Our findings confirm that USC testimonies tend to follow a more guided and structured format, especially in the early parts of the interview. Yale testimonies, by contrast, display greater topical fluidity, earlier emergence of core themes, and a higher proportion of open-ended questions. However, our analysis also uncovers significant convergences across both collections in later stages of the interviews, as well as similar narrative arcs, which seem to follow the chronology of Holocaust experience.

This article thus makes a dual contribution: (1) it offers a data-driven reassessment of a central historiographical claim within Holocaust testimony scholarship; (2) it introduces a scalable framework for computational comparison of large oral history corpora. By demonstrating how complex, ethically sensitive narratives can be analyzed computationally without compromising their integrity and interpretive richness, we open a pathway for further digital humanities work in trauma studies, oral history, and memory research.


\section{Oral history archives of the Holocaust}

\begin{figure*}[!ht] 
    \centering 
    \includegraphics[width=\textwidth]{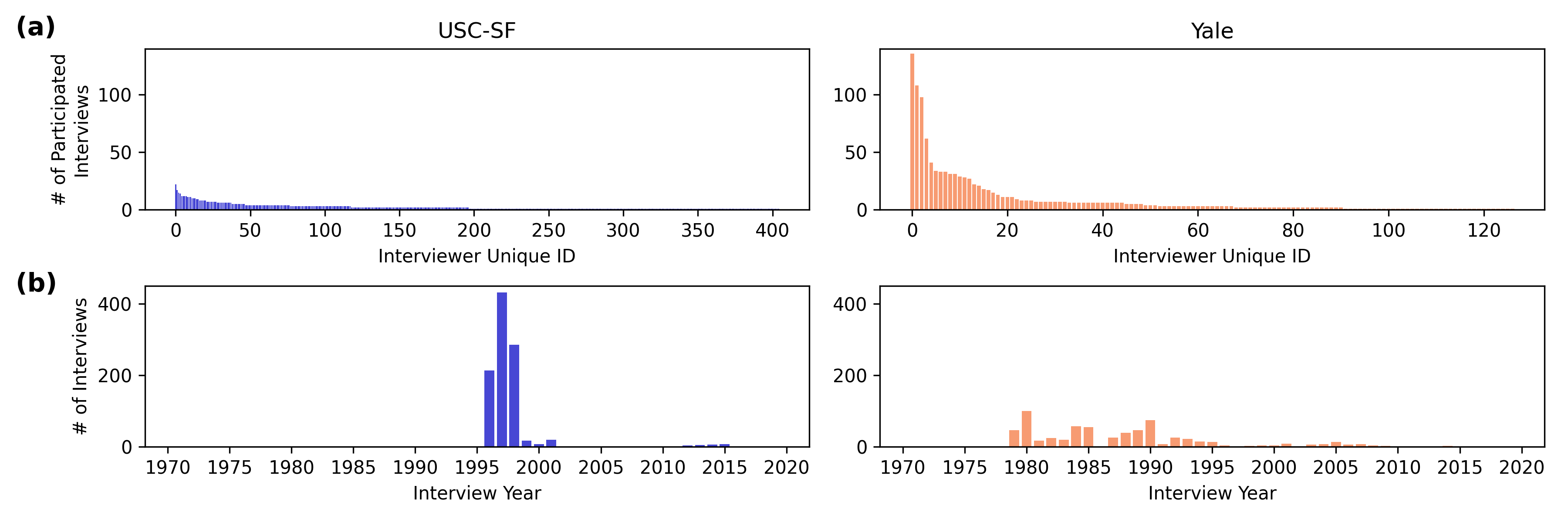} 
    \caption{(a) Number of interviews conducted (individually or jointly) by each interviewer in each archive. (b) Annual distribution of testimonies recorded across both archives, computed over the subset of testimonies available for this study.} 
    \label{fig:metadata} 
\end{figure*}

The Yale Fortunoff Video Archive for Holocaust Testimonies and the USC Shoah Foundation Visual History Archive are the two main and largest collections of video testimonies of Holocaust survivors. The Fortunoff Archive began in 1979 as a local initiative in New Haven, Connecticut, later relocating to Yale University \citep{hartman1995learning}. Holding more than 4,400 testimonies collected in the US, Europe, and Israel, its mission is to collect, preserve, and share the stories of those who were there. Being the first of its kind, the archive set the standard for projects to come \citep{felman1992testimony, hartman1996longest}. The USC archive was founded in 1994 by Steven Spielberg with the goal of recording 50,000 testimonies, for which hundreds of interviewers were trained in multiple countries \citep{smith2022trajectory}. It has since surpassed that goal, and in addition to Holocaust testimonies, has expanded to record victim narratives of other genocides (including Armenian, Rwandan, Cambodian) \citep{assmann2011cultural, smith2022trajectory}. Both archives were initially based on videotape technology and, over time, came to embrace digital technology, while incorporating and developing new methods to access and engage with testimonies \citep{assmann2011cultural, shandler2022digitizing}.

At its core, an oral history interview is a process in which an interviewer and an interviewee spend extended time together engaged in storytelling and attentive listening. It is a collaborative act of constructing a narrative. Unlike other forms of qualitative interviewing (such as in-depth interviews), oral history interviews tend to be less narrowly focused in their topical organization \citep{hesse_biber_practice_2005}. As with other oral history initiatives, each of the Holocaust archives has developed its own interview methodology.

From the outset, interviews recorded at Yale followed a clear code of conduct: putting the witnesses at the center and allowing them to relate their story freely with minimal guidance from interviewers. The interviewers’ main role was to provide survivors with a supportive listening companion during the interviews, all of which took place on the Yale campus \citep{felman1992testimony}. Conversely, USC interviews take place at the witness’s home and are typically preceded by a preparation meeting. USC employs a large number of interviewers in many countries, who all go through formal training and follow a set of elaborate and formulated guidelines \citep{uscshoah_interviewer_guidelines_2021}
\citep{Smith2022TrajectoryHolocaustMemory}). The Yale archive has relied on a relatively small group of interviewers, mostly academics, some with psychological training. As will be shown below, despite these disparities, interviews recorded by both archives tend to unfold chronologically, beginning with life before the war, through early signs of change, antisemitism and persecution, experiences during the war (ghetto, captivity, forced labor, extermination), end of war, and finally life after the war. All interviews examined here were transcribed manually by their respective archive, together with speaker identification and speaker-side segmentation.

It is hard to imagine contemporary Holocaust history and memory without the prevalence of survivors’ testimonies. The two archives have contributed significantly to the legitimacy of personal narratives as a source for the study of recent history, particularly of traumatic events \citep{Friedlander1993MemoryHistoryExtermination, hartman1996longest, Shandler2017HolocaustMemoryDigitalAge}.
By their nature, testimonies are based on personal memory whose accuracy might vary, but what they nevertheless express is something beyond factual knowledge about past events. It is the relating of the personal experience of how it felt, which expresses the human aspect often missing from grand historical accounts \citep{felman_laub_1992, langer1993holocaust}. As such, these narratives of pain and loss demand care and respect while retaining the singularity and integrity of each voice. 
Over the years, numerous audiovisual interviews with survivors have been recorded.
The practical impossibility of watching tens of thousands of such testimonies presents an unprecedented challenge for a morally informed study of recent history. It is precisely here that carefully designed digital tools can enable new, previously unattainable modes of engagement with testimony \citep{Keydar2020ListeningFromAfar,Keydar2022ChangingLensSurvivorTestimony}. 
Recent advances in natural language processing and large language models have opened new algorithmic ways of engaging with thousands of Holocaust testimonies at scale \citep{blanke2019understanding, naron2020let, ezeani2024geography, presner2024ethics, shizgal2025computational, keydar2026testimony}. 
However, to the best of our knowledge, there has so far been no attempt to harness these tools for large-scale comparative analysis across oral history archives.

\section{Data and Methodology}

This study analyzes 1,668 Holocaust survivor testimonies: 1,000 from the USC Shoah Foundation archive and 668 from the Yale Fortunoff Archive. Each interview was manually transcribed, annotated with speaker labels, and segmented into question–answer (Q/A) pairs, as outlined below. Due to the sensitive nature of these collections, transcripts are accessible only by direct request to the respective institutions. All testimonies in this study were conducted in English.

\subsection{Date of testimonies} \label{sec:data_date}

The two archives developed and followed highly different collection strategies. The Fortunoff Video Archive, established in 1979, has followed an open-ended acquisition policy, resulting in interviews collected over more than four decades. In contrast, the USC Shoah Foundation conducted a large-scale collection effort primarily between 1994 and 1999. However, within our available dataset, the testimonies from both archives are concentrated towards the initial years of operation. Fig. \ref{fig:metadata}(b) shows the distribution of testimonies per year in the subset of testimonies available for this study. 

\subsection{Interview Length} \label{sec:data_len}

USC interviews average 23,396 words\footnote{A word defined as any whitespace-separated token.} ($\sigma = 10{,}397$), while Fortunoff interviews average 13,622 words ($\sigma = 7{,}649$). Although both archives use similar methodologies, their protocols differ. One notable outcome is the substantial difference in length: USC interviews are, on average, 1.7 times longer than those from Fortunoff. This results in a greater quantity of content and potentially richer detail in the USC interviews. To mitigate the influence of this disparity, our quantitative metrics normalize for testimony length.  

\subsection{Distribution of interviewers} \label{sec:data_dist}

\begin{table}[ht]
\centering
\resizebox{\columnwidth}{!}{%
\small
\begin{tabular}{|c|l|c|l|c|}
\hline
 & \multicolumn{2}{|c|}{\textbf{USC}} & \multicolumn{2}{c|}{\textbf{Yale}} \\
\hline
\small{Top} & \small{Name} & \small{\#} & \small{Name} & \small{\#} \\
\hline
1   & Lorrie Fein           & 22 & Vlock Laurel             & 136 \\
2   & Esther Finder         & 17 & Laub Dori                & 108 \\
3   & Joanna Buchan         & 15 & Kline Dana               & 98 \\
4   & Reuben Zylberszpic    & 14 & Rudof Weiner             & 62 \\
5   & Joseph Huttler        & 12 & Millen Susan             & 41 \\
6   & Dina Cohen            & 12 & Frances Cohen            & 34 \\
7   & Ruth Meyer            & 12 & Langer Lawrence          & 33 \\
8   & Zepporah Glass        & 12 & Herz Moss                & 33 \\
9   & Florence Shuster      & 11 & Strochlic Kathy          & 31 \\
10  & Yvonne Walter         & 11 & Katz Helen               & 31 \\
\hline
\end{tabular}%
}
\caption{Top 10 participating interviewers and corresponding number of interviews conducted by the interviewer either as an individual or as part of a joint interview.}
\label{tab:top_interviewers}
\end{table}

Fig. \ref{fig:metadata}(a) shows the number of interviews each interviewer conducted in each archive. The Fortunoff corpus engaged a small cohort of interviewers, some of whom conducted over 100 interviews each, reflecting an approach that values a more personal interviewing style. In contrast, no individual interviewer in the USC corpus conducted more than 22 interviews (see top 10 participating interviewers in table \ref{tab:top_interviewers}). This distributed practice aligns with USC archive’s emphasis on standardized interview protocols, which prioritize methodological consistency over personal interviewing style. 

\subsection{Methodology}

Each testimony was segmented into chronological subunits using two complementary strategies, designed to capture both micro- and macro-level structures.

At the micro-level, we employed a topic-oriented segmentation: each Q/A pair was treated as a discrete topical unit, under the assumption (following prior work, \citep{ifergan2024identifying}) that individual exchanges typically cover single topics. Short Q/A pairs were merged with adjacent exchanges when appropriate to maintain contextual coherence. This segmentation allowed us to examine the dialogic aspects of interviewer–survivor interaction while retaining the structure of the narrative.

For macro-level analysis, we divided each testimony into $k=15$ equal-length chronological segments based on prior work (\citep{ifergan2024identifying}) and manual validation, enabling a normalized temporal comparison across testimonies of varying lengths and numbers of Q/A pairs. From each segment, we extracted a range of features such as topical diversity, segmental coherence, answer and question lengths, and question type classification using custom pipelines and GPT-based classification.

Thus, in this study, we quantify the differences in the ``structuredness'' across collections, emphasizing the topical and interviewer–survivor dialogic dynamics. Nonetheless, additional dimensions that may confound a comparative analysis between Yale and USC interviews are discussed in \S\ref{sec:data_date} - \S\ref{sec:data_dist}.


\begin{table*}[!ht]
\centering
\resizebox{\textwidth}{!}{%
\begin{tabular}{|c|c|c|c|}
\hline

\multicolumn{4}{|c|}{\cellcolor{colorLightGray}\textbf{USC-SF}} \\
\hline

\textbf{\cellcolor{colorDarkGray}Seg.} & 
\textbf{\cellcolor{colorDarkGray}Topic 1} &
\textbf{\cellcolor{colorDarkGray}Topic 2} &
\textbf{\cellcolor{colorDarkGray}Topic 3} \\
\hline

\cellcolor{colorDarkGray}1 &
\cellcolor{colorLightBlue}Childhood Memories (0.82)  &
\cellcolor{colorLightBlue}Family Heritage (0.87)  &
\cellcolor{colorLightPurple}Jewish Identity (0.67) \\
\hline

\cellcolor{colorDarkGray}2  &
\cellcolor{colorLightBlue}Childhood Memories (0.88)  &
\cellcolor{colorDarkPurple}Experiences of Antisemitism (0.48)  &
\cellcolor{colorLightBlue}Family Dynamics (0.76) \\
\hline

\cellcolor{colorDarkGray}3  &
\cellcolor{colorLightBlue}Childhood Memories (0.86)  &
\cellcolor{colorDarkPurple}Experiences of Antisemitism (0.61)  &
\cellcolor{colorLightOrange}Life in the Ghetto (0.36) \\
\hline

\cellcolor{colorDarkGray}4 &
\cellcolor{colorLightOrange}Life in the Ghetto (0.45) &
\cellcolor{colorMediumBlue}Family Separation (0.45) &
\cellcolor{colorBrown}Survival Strategies (0.69) \\
\hline

\cellcolor{colorDarkGray}5 &
\cellcolor{colorLightOrange}Life in the Ghetto (0.48) &
\cellcolor{colorMediumBlue}Family Separation (0.52) &
\cellcolor{colorBrown}Survival Strategies (0.74) \\
\hline

\cellcolor{colorDarkGray}6 &
\cellcolor{colorLightOrange}Life in the Ghetto (0.51) &
\cellcolor{colorBrown}Survival Strategies (0.76) &
\cellcolor{colorMediumBlue}Family Separation (0.53) \\
\hline

\cellcolor{colorDarkGray}7 &
\cellcolor{colorMediumBlue}Family Separation (0.51) &
\cellcolor{colorDarkOrange}Survival in Hiding (0.56) &
\cellcolor{colorLightOrange}Life in the Ghetto (0.52) \\
\hline

\cellcolor{colorDarkGray}8 &
\cellcolor{colorDarkOrange}Survival in Hiding (0.56) &
\cellcolor{colorMediumBlue}Family Separation (0.51) &
\cellcolor{colorLightOrange}Experiences in Auschwitz (0.33) \\
\hline

\cellcolor{colorDarkGray}9 &
\cellcolor{colorDarkOrange}Survival in Hiding (0.58) &
\cellcolor{colorMediumOrange}Life in Concentration Camps (0.52) &
\cellcolor{colorMediumBlue}Family Separation (0.47) \\
\hline

\cellcolor{colorDarkGray}10 &
\cellcolor{colorDarkOrange}Survival in Hiding (0.56) &
\cellcolor{colorMediumBlue}Family Separation (0.46) &
\cellcolor{colorYellow}Post-War Resilience (0.56) \\
\hline

\cellcolor{colorDarkGray}11 &
\cellcolor{colorDarkOrange}Survival in Hiding (0.52) &
\cellcolor{colorBlue}Family Loss and Reunion (0.40) &
\cellcolor{colorYellow}Post-War Identity Struggles (0.54) \\
\hline

\cellcolor{colorDarkGray}12 &
\cellcolor{colorBrown}Survival and Resilience (0.88) &
\cellcolor{colorBlue}Family Loss and Reunion (0.46) &
\cellcolor{colorYellow}Post-War Identity Struggles (0.64) \\
\hline

\cellcolor{colorDarkGray}13 &
\cellcolor{colorBlue}Family Loss in the Holocaust (0.44) &
\cellcolor{colorBrown}Survival and Resilience (0.85) &
\cellcolor{colorYellow}Post-War Identity Struggles (0.70) \\
\hline

\cellcolor{colorDarkGray}14 &
\cellcolor{colorDarkBlue}Family Memories (0.76) &
\cellcolor{colorLightGreen}Survival Reflections (0.79) &
\cellcolor{colorDarkGreen}Holocaust Legacy (0.69) \\
\hline

\cellcolor{colorDarkGray}15 &
\cellcolor{colorDarkBlue}Family Memories (0.82) &
\cellcolor{colorLightGreen}Legacy of Survival (0.78) &
\cellcolor{colorDarkGreen}Holocaust Remembrance (0.63) \\
\hline

\hline


\multicolumn{4}{|c|}{\cellcolor{colorLightGray}\textbf{Yale}} \\
\hline

\textbf{\cellcolor{colorDarkGray}Seg.} & \textbf{\cellcolor{colorDarkGray}Topic 1} & \textbf{\cellcolor{colorDarkGray}Topic 2} & \textbf{\cellcolor{colorDarkGray}Topic 3} \\
\hline

\cellcolor{colorDarkGray}1 &
\cellcolor{colorLightBlue}Childhood Memories Before the War (0.76) &
\cellcolor{colorLightPurple}Jewish Identity and Anti-Semitism (0.80) &
\cellcolor{colorLightBlue}Family Life Before the Holocaust (0.72) \\
\hline

\cellcolor{colorDarkGray}2 &
\cellcolor{colorLightOrange}Survival in the Ghetto (0.55) &
\cellcolor{colorMediumBlue}Family Separation (0.50) &
\cellcolor{colorDarkPurple}Experiences of Anti-Semitism (0.76) \\
\hline

\cellcolor{colorDarkGray}3 &
\cellcolor{colorLightOrange}Survival in the Ghetto (0.57) &
\cellcolor{colorMediumBlue}Family Separation (0.53) &
\cellcolor{colorDarkOrange}Hiding from Persecution (0.54) \\
\hline

\cellcolor{colorDarkGray}4 &
\cellcolor{colorLightOrange}Survival in the Ghetto (0.59) &
\cellcolor{colorMediumBlue}Family Separation Trauma (0.60) &
\cellcolor{colorMediumOrange}Experiences in Auschwitz (0.28) \\
\hline

\cellcolor{colorDarkGray}5 &
\cellcolor{colorLightOrange}Survival in the Ghetto (0.64) &
\cellcolor{colorMediumBlue}Family Separation (0.49) &
\cellcolor{colorMediumOrange}Experiences in Auschwitz (0.32) \\
\hline

\cellcolor{colorDarkGray}6 &
\cellcolor{colorDarkOrange}Survival Strategies in Hiding (0.58) &
\cellcolor{colorMediumBlue}Family Separation and Loss (0.61) &
\cellcolor{colorMediumOrange}Experiences in Concentration Camps (0.60) \\
\hline

\cellcolor{colorDarkGray}7 &
\cellcolor{colorDarkOrange}Survival in Hiding (0.54) &
\cellcolor{colorMediumBlue}Family Separation (0.48) &
\cellcolor{colorLightOrange}Life in the Ghetto (0.50) \\
\hline

\cellcolor{colorDarkGray}8 &
\cellcolor{colorDarkOrange}Survival in Hiding (0.53) &
\cellcolor{colorMediumBlue}Family Separation (0.52) &
\cellcolor{colorMediumOrange}Life in Concentration Camps (0.58) \\
\hline

\cellcolor{colorDarkGray}9 &
\cellcolor{colorMediumOrange}Survival in Concentration Camps (0.60) &
\cellcolor{colorLightOrange}Life in the Ghetto (0.47) &
\cellcolor{colorMediumBlue}Family Separation and Loss (0.55) \\
\hline

\cellcolor{colorDarkGray}10 &
\cellcolor{colorDarkOrange}Survival in Hiding (0.56) &
\cellcolor{colorBlue}Family Loss and Trauma (0.68) &
\cellcolor{colorYellow}Post-War Resilience (0.57) \\
\hline

\cellcolor{colorDarkGray}11 &
\cellcolor{colorDarkOrange}Survival in Hiding (0.56) &
\cellcolor{colorBlue}Family Loss in the Holocaust (0.42) &
\cellcolor{colorMediumOrange}Experiences in Concentration Camps (0.67) \\
\hline

\cellcolor{colorDarkGray}12 &
\cellcolor{colorBrown}Survival Strategies (0.77) &
\cellcolor{colorBlue}Family Loss (0.43) &
\cellcolor{colorYellow}Post-War Resilience (0.64) \\
\hline

\cellcolor{colorDarkGray}13 &
\cellcolor{colorMediumOrange}Survival in Concentration Camps (0.53) &
\cellcolor{colorBlue}Family Loss and Trauma (0.71) &
\cellcolor{colorYellow}Post-War Resilience (0.65) \\
\hline

\cellcolor{colorDarkGray}14 &
\cellcolor{colorBrown}Survival and Trauma (0.89) &
\cellcolor{colorBlue}Family Loss and Memory (0.68) &
\cellcolor{colorLightPurple}Identity After Liberation (0.73) \\
\hline

\cellcolor{colorDarkGray}15 &
\cellcolor{colorBrown}Survival and Memory (0.74) &
\cellcolor{colorBlue}Family Loss and Resilience (0.55) &
\cellcolor{colorDarkGreen}Holocaust Education and Remembrance (0.53) \\
\hline

\end{tabular}%
}
\caption{Top three topics per chronological segment. Recurring topics are manually color-coded across segments using a consistent palette showing the transformation of themes over time. Coverage values for each topic are displayed alongside the labels.} 
\label{tab:topic_coverage}
\end{table*}

\section{Findings}

\subsection{Topical Sequence Analysis}

\begin{figure*}[!ht] 
    \centering 
    \includegraphics[width=0.8\textwidth]{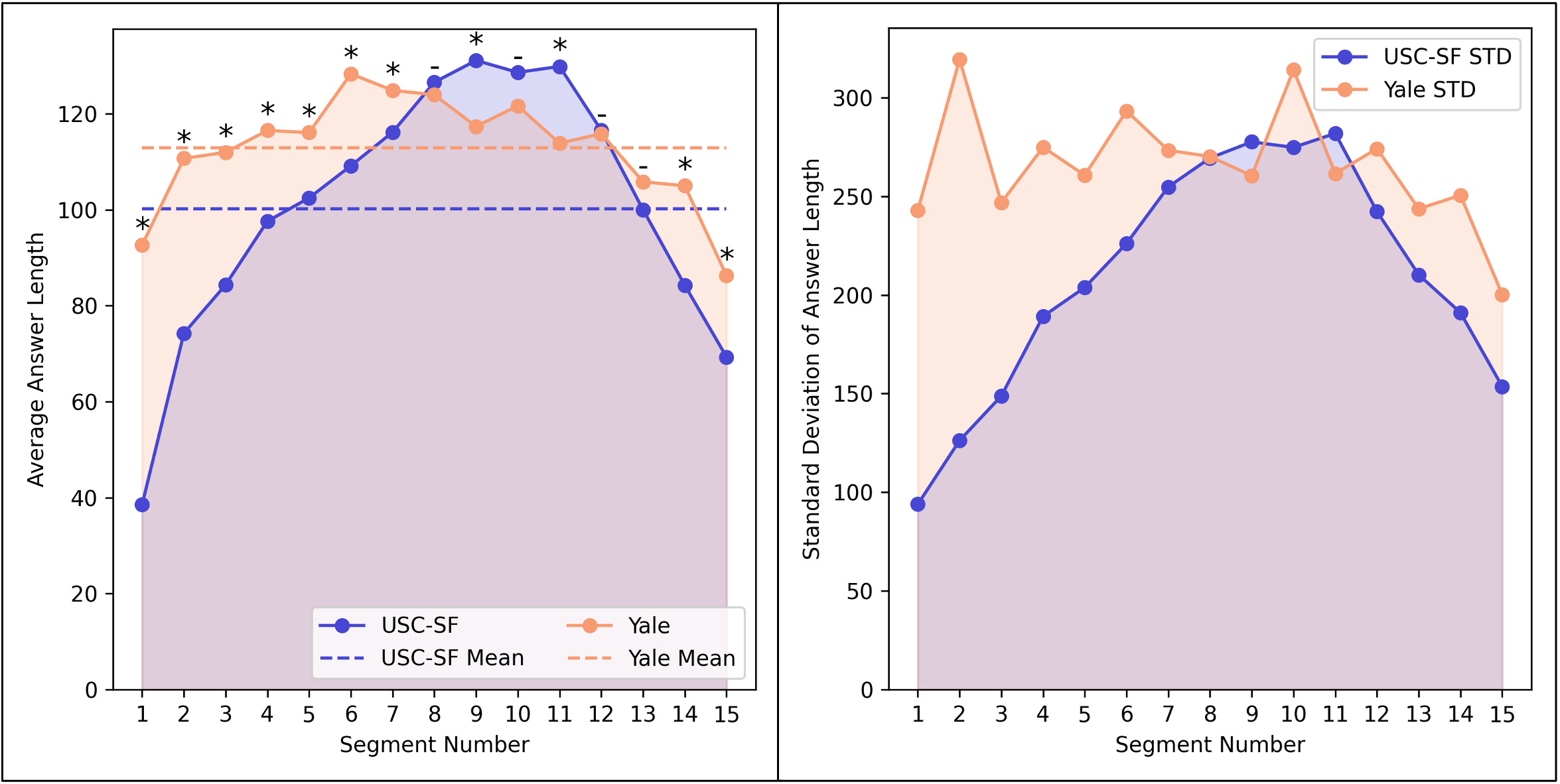} 
    \caption{Mean (left) and SD (right) of answer length over time for USC and Yale archives. Statistically significant ($t$-test) differences are marked with an asterisk (*).} 
    \label{fig:answer_length} 
\end{figure*}

\begin{figure*}[!ht] 
    \centering 
    \includegraphics[width=0.8\textwidth]{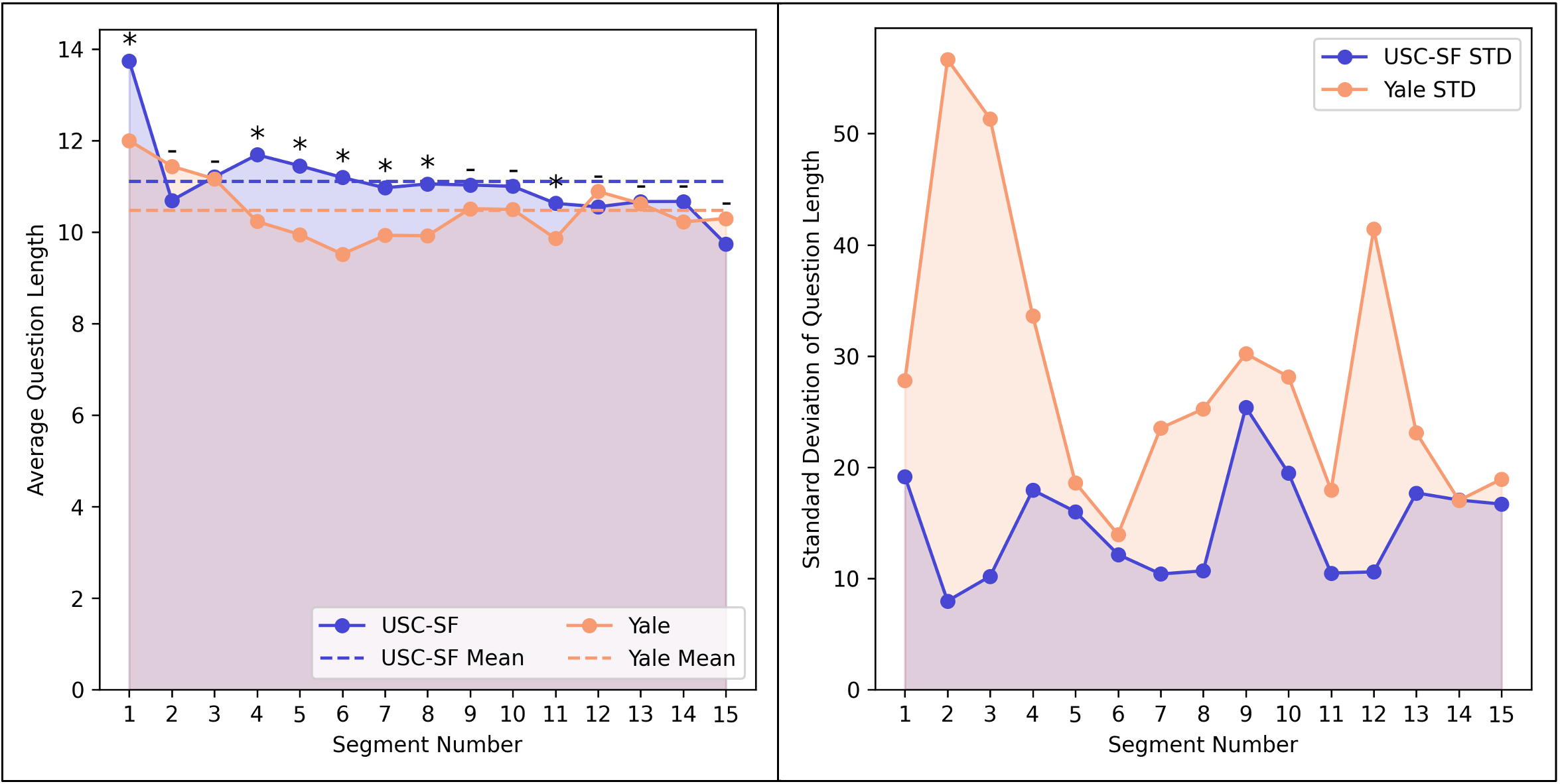} 
    \caption{Mean (left) and SD (right) of question length over time for USC and Yale archives. Statistically significant ($t$-test) differences are marked with an asterisk (*).} 
    \label{fig:question_length} 
\end{figure*}

A central dimension of ``structuredness'' in oral history interviews is the sequencing of topics across the testimonial timeline. Prior scholarship (e.g., \citep{piper2021narrative, wagner2022topical, ranade2022computational, ifergan2024identifying, wagner2025unsupervised, shizgal2025computational, trainin2025t5score}) suggests that highly structured interviews should exhibit a more predictable topical order, narrower chronological coverage per topic, and reduced thematic diversity across segments, whereas free-form interviews are expected to display looser sequencing and greater thematic fluidity.

To evaluate this hypothesis, we apply a computational topic-analysis pipeline modeled on, yet distinct from, earlier manual or semi-supervised approaches. We build upon the framework introduced in \citep{ifergan2024identifying}, but instead of employing topics generated by LDA \citep{blei2003latent} or BERTopic \citep{grootendorst2022bertopic} followed by expansive manual topic labeling and heuristic clustering thresholds, we introduce an alternative LLM-based strategy inspired by \citep{trainin2025t5score}. In doing so, we provide a scalable and fully automatic methodology for identifying emergent topical patterns across thousands of segments.

Our pipeline unfolds in two stages. First, an LLM (ChatGPT) generates a concise descriptive label for every Q/A pair, which serves as the micro-level unit of analysis (see Prompt \ref{prompt:topic_naming}). In the second stage (see Prompt \ref{prompt:common_topics}), we prompt the LLM to identify the Top-K recurring topics for each chronological segment. This staged approach captures both the local dynamics of interviewer–survivor exchanges and the broader macro-level narrative structure. Fig. \ref{fig:topic_diagram} in the appendix presents an illustration of this pipeline.

To validate our methodology, we calculated a Topic Coverage score \citep{trainin2025t5score} for each inferred topic using a random sample of 50 testimonies. Additionally, we compared the USC topics produced by our pipeline to those derived in \citep{ifergan2024identifying} and found strong structural convergence, indicating that the LLM-based approach is robust despite its limited interpretability.

Table \ref{tab:topic_coverage} visualizes the top three topics per chronological segment. Recurring topics are manually color-coded across segments using a consistent palette, enabling readers to trace the persistence or transformation of themes over time. Coverage values for each topic are displayed alongside the labels.

Across both corpora, we observe a clear chronological trajectory: testimonies move from prewar life, to the onset of persecution, to deportation and camp experiences, and ultimately to liberation and postwar recovery. One salient structural difference, however, is the USC archive’s recurring concluding segment dedicated to discussing personal photographs, labeled by our method as ``Family Memories''. This photo segment appears consistently in USC interviews but is largely absent from the Yale corpus, marking a distinct divergence in institutional interviewing protocols.

Despite the overall similarity of the macro-historical arc, the two corpora diverge in topical behavior and narrative dynamics. USC testimonies display a more predictable and segmented topical progression, with clearer boundaries between phases, an effect aligned with the archive’s structured interview guidelines. Yale testimonies, in contrast, exhibit greater thematic fluidity: topics begin earlier, overlap more frequently, and persist across multiple segments, producing a narrative rhythm that is less tightly scaffolded by interviewer intervention.

Patterns of topical diversity likewise differ. In Yale interviews, the dominant topics in early segments tend to be introspective and affective, with broader diversification emerging only by the third topic. USC interviews, by contrast, show greater topic variation as early as the second and third topics, yielding a more multidimensional narrative distribution. These differences reflect not only interviewing style but also the interactional norms cultivated by each institution.

Taken together, the results complicate the binary distinction between “structured” and “free-form” interviews. Both archives follow a broadly linear historical logic, yet they diverge in how topics arise, persist, and shift across the testimonial timeline. The Yale corpus tends toward emotional continuity and reflective narration, while the USC corpus exhibits a more segmented and guided progression. The combination of LLM-based topic extraction, cross-method validation (LDA, BERTopic), and explicit visualization reveals that ``structuredness'' in Holocaust testimony is not merely a function of protocol but a dynamic interplay between institutional design, interviewer choices, and survivor narrative agency.


\subsection{Question-Answer Dynamics}

To measure question and answer dynamics, we performed a simple word count using the \texttt{NLTK} \citep{bird-loper-2004-nltk} library. We then plotted changes in average word count across testimony segments for each corpus, reporting both the mean and variance. A two-sample $t$-test was conducted for each segment to assess whether differences between the corpora were statistically significant (indicated by an asterisk). The overall mean length was plotted as a dashed reference line. See Figures \ref{fig:answer_length}, \ref{fig:question_length}. 

\textbf{Answer Length}: Yale survivors tended to provide longer answers in the early segments of their testimonies, whereas responses in the USC corpus gradually lengthened over time. Approximately between segments 1–8, Yale testimonies exhibited longer responses on average. Around segments 8–11, this pattern reversed, with USC answers becoming longer. Toward the end of the testimonies, both collections showed a relative decline in answer length, although Yale participants generally maintained slightly longer responses.

Examining the standard deviation (SD) of answer length reveals the degree of variability across testimonies. Higher SD values indicate greater fluctuation in response length, which may reflect differences in survivors’ narrative styles, interviewer intervention, or emotional pacing. In this context, Yale testimonies exhibit higher variability, suggesting that survivors alternated between extended reflections and shorter responses, while USC testimonies show more consistent pacing and structure. 

\textbf{Question Length and Frequency}: USC interviewers tended to ask longer and more frequent questions, particularly during the early portions of the testimonies. This pattern aligns with the more guided and segmented structure of USC interviews. In contrast, Yale interviewers posed shorter and less predictable questions. The high variability in question length within the Yale corpus (as reflected in SD values) is noteworthy, given the relatively small pool of interviewers, suggesting considerable stylistic diversity and less standardized interviewing practice. 

\textbf{Convergence}: Both archives displayed a reduction in answer length toward the final segments, a trend likely associated with standardized closing routines or reflective conclusions. Statistical testing confirmed that differences in answer length between the two corpora were significant in the early interview stages but diminished progressively toward the end.

Taken together, the analysis of question and answer length reveals distinct narrative and methodological dynamics within the two archives. The longer early responses in the Yale corpus suggest a more open and reflective interview style, one that allows survivors to elaborate freely and set the narrative pace. In contrast, the USC interviews begin with shorter, more structured exchanges but expand over time, indicating an interviewer-driven scaffolding that gradually gives way to survivor-led narration. The higher variability in Yale question and answer lengths reflects more dialogical spontaneity, where interviewers adapt responsively to the survivor’s storytelling, whereas the more uniform USC patterns demonstrate adherence to a formalized interview guide. The convergence observed in later segments, marked by shorter responses and reduced differences between the corpora, points to the influence of shared testimonial conventions and closing rituals that shape the end of interviews across institutions. Overall, these patterns highlight that interview structure is not static but evolves through the testimony, balancing institutional design with survivor agency and emotional rhythm.

\begin{figure}[!ht]
    \centering 
    \includegraphics[width=\columnwidth]{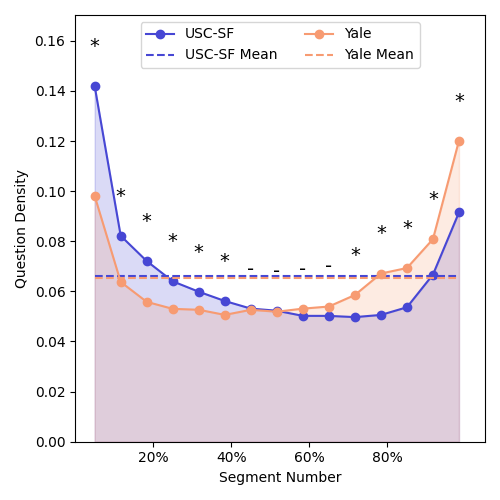} 
    \caption{Intervention Density for USC and Yale archives. Statistically significant ($t$-test) differences are marked with an asterisk (*).} 
    \label{fig:qa_density} 
\end{figure}

\subsection{Intervention Density}

To assess the density of interviewer interventions, we measured the duration of uninterrupted survivor speech. Lacking timestamp data, we used word count as a proxy for duration by calculating the proportion of uninterrupted survivor words relative to the total testimony length.

To examine how this measure changes over time, we divided each interview into 15 equal segments based on the cumulative word count. This segmentation, therefore, represents proportional divisions of the testimony length rather than the micro-level segmentation used in the Q/A-based analysis. For each corpus, we computed the average intervention density per segment across all testimonies. This word-based segmentation allows us to capture the relative distribution of interviewer interventions throughout the testimonies, independent of total interview length, and provides a comparable temporal framework across the two archives. The results, presented in Fig. \ref{fig:qa_density}, display the average uninterrupted speech density for each segment.

The analysis of intervention density provides further insight into the interactional dynamics that differentiate the two archives. Longer uninterrupted segments of survivor speech indicate greater narrative autonomy, whereas higher intervention density suggests tighter interviewer control or increased dialogical guidance. In this context, the Yale testimonies tend to feature longer uninterrupted stretches, reflecting a more survivor-centered approach that privileges emotional flow and narrative continuity. The USC interviews, by contrast, exhibit a denser pattern of interventions, especially in the earlier portions of the testimonies, which is consistent with a structured interview format emphasizing chronological order. However, in both collections, there is a gradual reduction in intervention density as testimonies progress, suggesting perhaps increased confidence on the survivors’ part or greater flexibility on the interviewers' part. This suggests that narrative authority is negotiated throughout the testimonial encounter, resulting in an evolving dynamic between institutional control and individual agency in the making of oral history.

\subsection{Question Types}

\begin{figure}[!ht] 
    \centering 
    \includegraphics[width=\columnwidth]{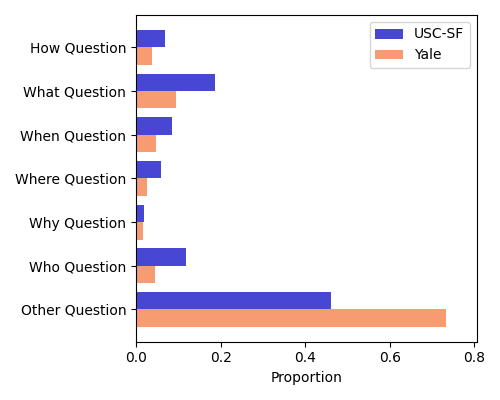} 
    \caption{Distribution of question types for USC and Yale archives.} 
    \label{fig:question_types} 
\end{figure}

Building on a framework proposed in (\citep{mittelstadt2016ethics}, Fig. 3.6), we hypothesize that interviewing style affects the distribution of question types–specifically “why,” “what,” “where,” “who,” “when,” “how,” and “other”. Structured interviews are expected to rely more heavily on factual and directive questions (e.g., “what,” “when,” “who”), whereas less structured or conversational interviews are likely to include a broader range of open-ended or ambiguous questions (classified as “other”).

\begin{figure*}[!ht]
    \centering 
    \includegraphics[width=0.8\textwidth]{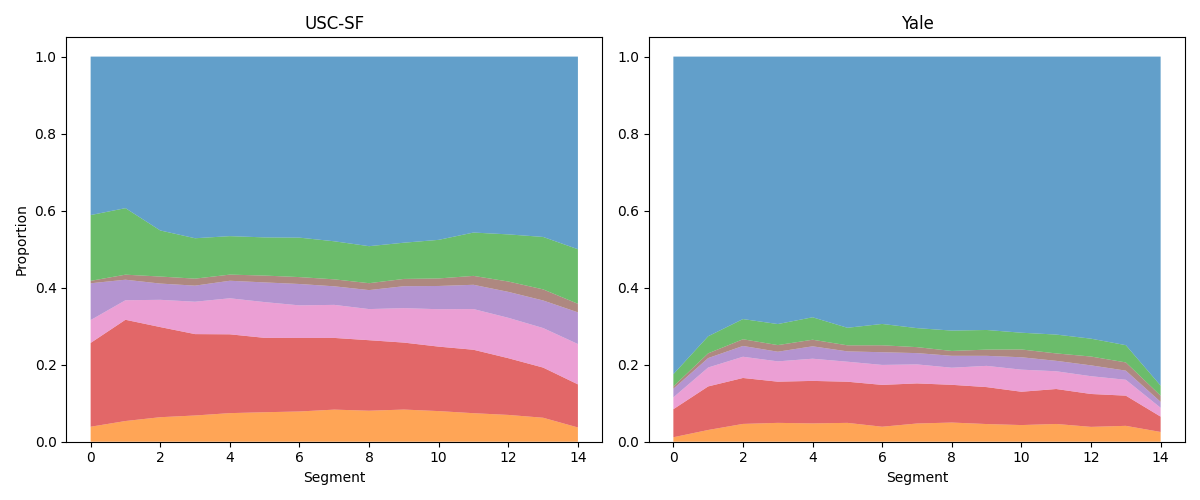} 
    \caption{Distribution of question types over time for USC and Yale archives.} 
    \label{fig:question_types_over_time} 
\end{figure*}

To test this hypothesis, we extracted all interviewer questions from both testimony corpora and employed an LLM to classify each into one of the seven categories (see Prompt \ref{prompt:q_type} in the Appendix). Model predictions were validated through manual review of a random subset of 50 examples per category (3,500 questions in total). The overall and segment-based distributions of question types are presented in Figures \ref{fig:question_types} and \ref{fig:question_types_over_time}.

The LLM-based classification revealed clear stylistic differences between the two archives. Yale interviewers used a higher proportion of “Other” questions, exhibiting a more flexible and exploratory interviewing style that allows survivors to guide the conversation. In contrast, USC interviewers relied predominantly on “what,” “who,” and “when” questions, particularly in the early segments, consistent with a more structured and documentary-oriented approach. That said, USC interviews show a gradual increase in “Other” question types over time, which might indicate that as rapport developed, the exchanges became more dialogic and open-ended.

Overall, the distribution of question types offers insight into institutional interviewing practices and the dynamics of testimonial co-construction. The USC format prioritizes a more formal interview structure, limiting opportunities for narrative divergence or emotional elaboration. The Yale format fosters greater conversational variability and affective depth, which allows far greater survivor agency in shaping the narrative, but for this reason is more dependent on survivors’ compliance and initiative. These findings illustrate how question type mediates the balance between institutional protocol and personal voice, revealing how methodological design and interpersonal rapport together shape the structure and tone of Holocaust testimony.


\section{Toward a Scalable Framework for Oral History Comparison}

Our analysis confirms that USC interviews are, overall, more structured than those in the Yale archive, particularly in question types, topical flow, and interviewer interventions. The USC corpus displays clearer segmental boundaries and a more predictable topical progression, whereas Yale testimonies reveal greater thematic fluidity, variability in questioning style, and more open-ended conversational turns. As the interview unfolds, however, these contrasts diminish as both interview styles move toward similar narrative rhythms shaped by survivor agency, historical chronology, and the affective dynamics of testimony. Ultimately, despite institutional differences, what becomes apparent is that in both collections the dialogical exchange between interviewer and survivor meaningfully shapes the unfolding of testimony alongside the structural template. This finding underscores that ``structuredness'' in oral history is not an institutional constant but a dynamic property of interaction, emerging from the interplay between methodological design, interpersonal trust, and the survivor’s narrative agency.

Beyond these findings, the study contributes a concrete methodological framework for comparative oral history analysis. The proposed pipeline introduces a sequence of computational strategies designed specifically for dialogic and large-scale testimonial data. 

First, it develops a segmentation method that extracts coherent question-answer trajectories from long interviews, enabling alignment across archives that differ in format and length. This segmentation serves as the foundation for cross-corpus comparison, allowing structural and thematic patterns to be analyzed at a shared level of granularity.

Second, the study advances a novel approach to LLM-based topic extraction. Working with oral history presents unique challenges, including contextual overflow, diffuse topic boundaries, and highly personal narrative structures. To address these, the pipeline employs a staged prompting strategy inspired by map-reduce logic: local prompts generate micro-level topics for each segment, and a secondary synthesis step aggregates these into common themes across the corpus. This approach mitigates the limitations of unsupervised topic modeling and allows for interpretable, reproducible results grounded in qualitative meaning.

Third, the framework integrates iterative evaluation and refinement. Through experimentation with classification strategies, segmentation thresholds, and prompt design, the study develops an ontology of best practices for LLM-assisted oral history analysis. The resulting workflow balances automation and interpretation, enabling scalable comparison while preserving the discursive and emotional complexity of survivor narratives.

Taken together, these contributions establish a reproducible and extensible model for comparing oral history archives at scale. Rather than treating computational methods as an abstraction from humanistic analysis, the pipeline operationalizes interpretive categories such as ``structuredness, topical coherence, and interviewer style into measurable and comparable features. It allows distinct institutional collections to be analyzed within a shared framework, revealing how methodological design, interview dynamics, and survivor agency jointly shape the production of testimony.

More broadly, this study demonstrates how digital humanities can move from isolated case studies toward comparative infrastructures that connect archives and traditions of memory work. The pipeline developed here enables not only new forms of analysis but also new questions about the ethics, scale, and dialogical nature of historical testimony. It shows that computational methods, when used critically, can deepen rather than diminish the interpretive work of the humanities, thus illuminating the ways in which stories are told, recorded, and remembered across time and institutional boundaries.

\section{Ethics Statement and Limitations}

Throughout the study, the OpenAI-GPT suite of models was used, with manual validation of segment samples conducted to confirm model consistency. Furthermore, all data access was obtained with institutional permission and handled strictly in accordance with ethical standards for trauma data. Future work will involve integrating this pipeline into an open-access annotation platform for cross-disciplinary use.




\section*{References}
\bibliographystyle{lrec2026-natbib}
\bibliography{reference.bib}


\onecolumn

\section*{Appendix}




\begin{figure*}[ht] 
    \centering 
    \includegraphics[width=\textwidth]{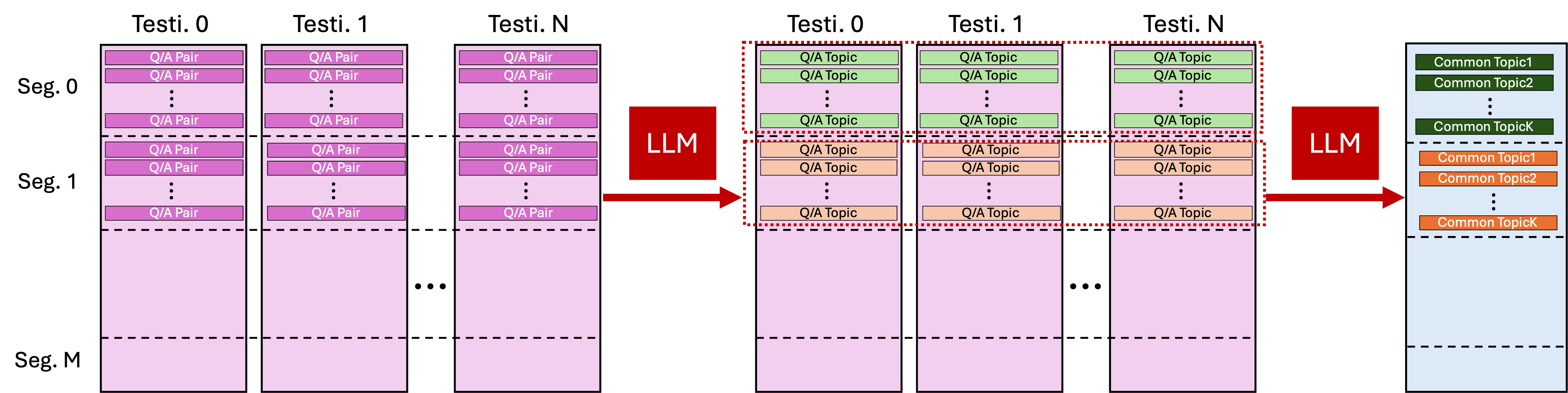} 
    \caption{The topical sequence extraction pipeline.} 
    \label{fig:topic_diagram} 
\end{figure*}

\section{LLM Prompts}

\subsection{Q/A Topic Naming} \label{prompt:topic_naming}





\begin{verbatim}
"""
You are a Holocaust researcher. 
You will be presented with a short text snippet from a 
conversation between an interviewer and a Holocaust survivor.
'INT' represents the interviewer and 'SUBJECT' or 
'<survivor_name>' represents the survivor.
Given the text snippet please generate a short title describing
the most prominent topic in the text.
Make sure that the title is short and limited to a few words.
Make sure that the title is comprehensive, specific, 
interpretable, and short.
Make sure that the title captures only a single topic.
Output format:
Title: "<title>"
Reason: "<reason>"
Text snippet:
"{text_snippet}"
"""
\end{verbatim}

\subsection{Common Topics Extraction} \label{prompt:common_topics}
\begin{verbatim}
"""
You are a Holocaust researcher. 
You will be presented with a set of titles representing topics
extracted from Holocaust survivor interviews.

Title Set:
{title_set}

Your task is:
- Generate {num_topics} distinct titles that best describe the
most common and prominent titles in set.
- Titles must be concise (maximum of a few words), specific,
interpretable, and distinct.
- Do NOT combine multiple topics into a single title or use
conjunctions like “and”.

Desired output format:
{output_format}

The common titles are:
1.
"""
\end{verbatim}

\subsection{Question Type Classification} \label{prompt:q_type}
\begin{verbatim}
"""
You are a Holocaust researcher analyzing survivor testimonies.
The testimonies are transcripts of an oral interviews.
The interviews is composed of speaker sides -- the interviewer
questions and the survivor answers. 
A prefix of "INT" identifies an "interviewer" line.
A prefix of "<initials>" identifies a suviror line.
You will be given one question asked by an interviewer during a
conversation with a Holocaust survivor.

Your task is to classify the question into exactly **one** of 
the following types, based on its structure and intent:

Question Types:
1. How Question – Asks about a method, process, or manner 
2. What Question – Asks for information, descriptions, or 
clarifications 
3. When Question – Asks about time or timing 
4. Where Question – Asks about a place or location 
5. Why Question – Asks about cause, motivation, or reason 
6. Who Question – Asks about a person or group 
7. Other Question – Does not match any of the categories 
above or is ambiguous.

Then, write a brief explanation (no more than one sentence) 
explaining why you selected that type.

Do not return more than one type.

### Output JSON format:
{{
    "type": "<one of the 7 types above>",
    "explanation": "<one-sentence explanation>"
}}
    
### Input:
Interviewer Question: "{speaker_line}"
"""
\end{verbatim}

\end{document}